\documentclass{article}

\PassOptionsToPackage{table}{xcolor}

\usepackage[utf8]{inputenc} 
\usepackage[T1]{fontenc}    
\usepackage{microtype}      
\usepackage{graphicx}       
\usepackage{booktabs}       
\usepackage{multirow}       
\usepackage{amsmath}        
\usepackage{amssymb}        
\usepackage{amsfonts}       
\usepackage{nicefrac}       
\usepackage{xcolor}         
\usepackage{hyperref}       

\usepackage[accepted]{icml2026}
\usepackage[abbreviations]{glossaries-extra}
\setabbreviationstyle{long-short}
\glsdisablehyper
\usepackage{listings}
\lstdefinestyle{arxivpython}{
  language=Python,
  basicstyle=\ttfamily\scriptsize,
  keywordstyle=\color{blue!60!black},
  commentstyle=\color{green!40!black},
  stringstyle=\color{red!55!black},
  frame=lines,
  breaklines=true,
  breakatwhitespace=false,
  columns=fullflexible,
  keepspaces=true,
  showstringspaces=false,
  tabsize=4,
  captionpos=b
}

\icmltitlerunning{VASAE: Naming SAE Dictionary Directions}

\newabbreviation[longplural={Sparse Autoencoders}, shortplural={SAEs}]{sae}{SAE}{Sparse Autoencoder}
\newabbreviation{vasae}{VASAE}{Vocabulary-Aligned Sparse Autoencoder}
\newabbreviation[longplural={Large Language Models}, shortplural={LLMs}]{llm}{LLM}{Large Language Model}
\newabbreviation{pca}{PCA}{principal component analysis}
\newabbreviation[longplural={multi-layer perceptrons}, shortplural={MLPs}]{mlp}{MLP}{multi-layer perceptron}
\newabbreviation{ve}{VE}{variance explained}
\newabbreviation{ce}{CE}{cross-entropy}
\newabbreviation{mse}{MSE}{mean-squared error}
\newabbreviation{ioi}{IOI}{Indirect Object Identification}
\newabbreviation{kldiv}{KL}{Kullback--Leibler divergence}
\newabbreviation{pos}{POS}{part-of-speech}
\newabbreviation[longplural={Institutional Review Boards}, shortplural={IRBs}]{irb}{IRB}{Institutional Review Board}
\newabbreviation{gpt2}{GPT-2}{Generative Pre-trained Transformer 2}
\newglossaryentry{logitlens}{
	name={LogitLens},
	text={LogitLens},
	first={LogitLens},
	description={A method that projects intermediate post-residual streams through the unembedding matrix to inspect token predictions}
}

\newcommand{\R}{\mathbb{R}}              

\DeclareMathOperator*{\argmax}{arg\,max}

\DeclareMathOperator{\relu}{ReLU}

\DeclareMathOperator{\MLP}{MLP}
\DeclareMathOperator{\Attn}{Attn}

\DeclareMathOperator{\CE}{CE}

\DeclareMathOperator{\TopK}{TopK}

\newcommand{\norm}[1]{\left\lVert #1 \right\rVert}

\newcommand{\loss}{\mathcal{L}}          

\newcommand{\layer}{\ell}                

\newcommand{\vocab}{\mathcal{V}}         

\newcommand{\WE}{\mathbf{W}_{E}}         
\newcommand{\WU}{\mathbf{W}_{U}}         

\newcommand{\h}{\mathbf{h}}            

\newcommand{\logits}{\mathbf{u}}         
\newcommand{\probv}{\mathbf{p}}          

\newcommand{\dmodel}{d_{\mathrm{model}}}

\newcommand{\nlayer}{L}

\newcommand{\dvocab}{|\mathcal{V}|}
\newcommand{\dsparse}{d_\mathrm{sparse}}

\newcommand{\saecode}{\mathbf{z}}        
\newcommand{\topk}{top-$k$}              
\newcommand{\saerecon}{\tilde{\mathbf{h}}} 
\newcommand{\enc}{\mathcal{E}}
\newcommand{\dec}{\mathcal{D}}
\newcommand{\Wenc}{\mathbf{W}_{\enc}}
\newcommand{\Wdec}{\mathbf{W}_{\dec}}
\newcommand{\benc}{\mathbf{b}_{\enc}}
\newcommand{\bdec}{\mathbf{b}_{\dec}}

\newcommand{\secref}[1]{Section~\ref{#1}}
\newcommand{\subsecref}[1]{Subsection~\ref{#1}}
\newcommand{\appref}[1]{Appendix~\ref{#1}}
\newcommand{\figref}[1]{Figure~\ref{#1}}
\newcommand{\tabref}[1]{Table~\ref{#1}}
\newcommand{\eqnref}[1]{Equation~\ref{#1}}

\begin{document}

\twocolumn[
  \icmltitle{VASAE: Naming SAE Dictionary Directions with Vocabulary-Aligned Anchoring}

  \begin{icmlauthorlist}
    \icmlauthor{Kairui Zhang}{bristol-is}
    \icmlauthor{Ziwen Yu}{bristol}
    \icmlauthor{Zahraa S. Abdallah}{bristol-is}
    \icmlauthor{Martha Lewis}{uva}
  \end{icmlauthorlist}

  \icmlaffiliation{bristol-is}{Intelligent Systems Laboratory, University of Bristol, Bristol, UK}
  \icmlaffiliation{bristol}{University of Bristol, Bristol, UK}
  \icmlaffiliation{uva}{University of Amsterdam, Amsterdam, Netherlands}

  \icmlcorrespondingauthor{Kairui Zhang}{pu22650@bristol.ac.uk}
  \icmlkeywords{Sparse Autoencoders, Mechanistic Interpretability, Vocabulary Alignment}

  \vskip 0.3in
]

\printAffiliationsAndNotice{}

\begin{abstract}
  Sparse autoencoders (SAEs) provide useful decompositions of Transformer residual streams, but their learned features are usually named post hoc rather than directly connected to the Transformer's token vocabulary. We introduce Vocabulary-Aligned Sparse Autoencoder (VASAE), a method that trains SAE features under vocabulary-aligned anchoring and assigns each feature an intrinsic token name: the token string whose embedding is nearest to that feature. Without reducing reconstruction quality compared with a standard SAE, VASAE produces dictionaries with vocabulary-aligned features. Using a 0.8 cutoff on the nearest-token alignment score, dictionaries trained on GPT-2-small post-residual streams align about 90\% of features in layers 0--10. In Llama-3.1-8B, representative shallow and middle-layer dictionaries contain strongly aligned features, including 92.8\% in the shallow layer, while the representative final-layer dictionary shows limited alignment. After subtracting the sentence-level mean sparse code, case studies show that many remaining intrinsic token names are relevant to nearby input tokens. These results suggest that vocabulary-aligned anchoring can connect learned features to intrinsic token names during training, complementing post hoc interpretation of learned dictionaries.
\end{abstract}

\begin{center}
\small Project page: \url{https://karry-z.github.io/VASAE/}
\end{center}

\glsresetall

\section{Introduction}

Transformer architectures are a central model class for generative AI~\cite{vaswaniAttentionAllYou2017a,brownLanguageModelsAre2020a}, and language-model performance has improved with model and data scale~\cite{hoffmannEmpiricalAnalysisComputeoptimal2022}. Their behavior, however, is produced by high-dimensional internal states that are difficult to inspect directly~\cite{belinkovAnalysisMethodsNeural2019,ghandehariounPatchscopesUnifyingFramework2024}. In decoder-only Transformers, one important state is the \emph{residual stream}: the vector channel that each Transformer block reads from and writes to~\cite{elhage2021mathematical}. Understanding this stream matters because it carries the intermediate information from which later layers compute next-token predictions~\cite{gevaTransformerFeedForwardLayers2021a,ghandehariounPatchscopesUnifyingFramework2024}. 

\Glspl{sae} are increasingly used to decompose residual streams into sparse codes over learned dictionaries~\cite{hubenSparseAutoencodersFind2024,gaoScalingEvaluatingSparse2025}. In this paper, the dictionary is the decoder weight matrix, and a \emph{feature} is one column, or direction, of that matrix. A \emph{sparse code} is a vector of feature coefficients. Each entry corresponds to one feature. For a given residual-stream vector, most entries are zero, and a larger nonzero value means that the corresponding feature contributes more strongly to the reconstruction. Recent work uses \gls{sae} features for circuit discovery as interpretability tools~\cite{marksSparseFeatureCircuits2025,DBLP:conf/icml/KarvonenRLTBCLF25}. Standard \glspl{sae}, however, learn features for reconstruction and sparsity rather than direct naming~\cite{hubenSparseAutoencodersFind2024,gaoScalingEvaluatingSparse2025}. Feature labels are usually assigned after training by inspecting contexts where the corresponding element of the sparse code is large, or by running a separate automated interpretation procedure~\cite{bills2023language,pauloAutomaticallyInterpretingMillions2025}, making interpretation a separate post-hoc step. \Gls{vasae} assigns a training-time nearest-token label to each learned decoder direction.

The technical bottleneck is that the dictionary geometry and the vocabulary geometry are usually disconnected. Prior work shows that language-model representation spaces have nontrivial geometric structure, including anisotropy and dominant directions~\cite{ethayarajhHowContextualAre2019a,muAllbuttheTopSimpleEffective2018a}. Separately, weight-tying work shows that input and output embeddings are a central vocabulary interface in language models~\cite{pressUsingOutputEmbedding2017,inanTyingWordVectors2017}. A learned feature can reconstruct activations while lying far from any token embedding, so it has no intrinsic vocabulary-level name. The opposite design would force the decoder to equal the token embedding matrix; in this paper we test whether this hard-tied decoder is sufficient. The missing object is therefore a learnable dictionary that remains flexible for reconstruction while staying close enough to vocabulary directions to provide candidate token-based names. A token name is a geometry-derived label for a direction.

Our key observation is that the language model already contains a model-internal, vocabulary-indexed set of reference directions when the token embeddings share dimensionality with the SAE decoder space: the fixed input token embeddings~\cite{vaswaniAttentionAllYou2017a}. We use these embeddings as anchors, not as frozen features. Based on this observation, we propose \gls{vasae}, a \gls{sae} whose learned features are trained with the usual sparse reconstruction objective plus a vocabulary-aligned anchor objective. Each feature is encouraged to stay close to its nearest token embedding, and after training receives an \emph{intrinsic token name}: the token string whose embedding is nearest to that feature. Here intrinsic means that the name is induced by the trained feature's location relative to the model's own vocabulary embeddings.

Empirically, \gls{vasae} achieves reconstruction metrics comparable to a standard \gls{sae} under the tested settings while adding a geometric vocabulary-alignment signal. Across layers, it has comparable \gls{ve} on GPT-2-small ($0.965$) and Llama-3.1-8B ($0.931$), while hard-tying the decoder to token embeddings has lower reconstruction metrics. On GPT-2-small, $89$--$94\%$ of features in layers $L0$--$L10$ exceed the diagnostic alignment cutoff. With a larger tested anchor coefficient ($\lambda_{\text{anchor}}=5\times10^{-3}$), Llama-3.1-8B reaches $92.8\%$ at $L0$ but remains unstable in the final representative layer. Case studies show intrinsic token names assigned to active features in context.

Our contributions are:
\begin{itemize}
  \item We introduce \gls{vasae}: a soft vocabulary-anchoring objective that assigns learned \gls{sae} decoder directions nearest-token names.
  \item We evaluate reconstruction under vocabulary anchoring and test whether hard-tying the decoder to token embeddings is sufficient.
  \item We analyze embedding-space feature alignment and vocabulary coverage, and use case studies to show intrinsic token names as geometry-derived identifiers.
\end{itemize}

\section{Background}

Residual-stream \glspl{sae}, token embeddings, and vocabulary readouts all depend on the geometry of Transformer residual states. \gls{vasae} uses this geometry to define vocabulary-indexed anchors for learned \gls{sae} decoder directions.

\subsection{Transformer Residual Stream}

Transformer models maintain a persistent residual stream, which carries information across layers~\cite{vaswaniAttentionAllYou2017a,elhage2021mathematical}. Panel A of \figref{fig:arch} shows the standard Transformer structure used in this setup. Let $\vocab$ be the vocabulary space and let $\WE \in \R^{\dvocab \times \dmodel}$ be the token embedding matrix, where $\dmodel$ is the model dimension. Throughout the paper, we use a row-vector convention to stay consistent with implementations. For a token $v\in\vocab$, we denote its embedding by $\mathbf{w}_v\in\R^{\dmodel}$ and write the initial post-residual stream as $\h^{(0)}=\mathbf{w}_v$.

At each layer $\layer$, the residual stream is updated by adding the outputs of the attention and \gls{mlp} modules:
\begin{equation}
  \h^{(\layer)} = \h^{(\layer-1)} + \Attn_{\layer}(\h^{(\layer-1)}) + \MLP_{\layer}(\h^{(\layer-1)}).
\end{equation}
This schematic update omits the sequential sublayer structure and LayerNorm operations of the underlying Pre-LN Transformer blocks.

We call $\h^{(\layer)} \in \R^{\dmodel}$ the layer-$\layer$ post-residual stream: the residual stream after layer $\layer$ has added its attention and \gls{mlp} outputs. In this work, all \gls{sae} variants are trained and evaluated on the post-residual stream, not on the intermediate attention or \gls{mlp} outputs before they are added back into the residual stream.

For autoregressive language models, the final prediction logits are obtained by projecting the last-layer post-residual stream through the unembedding matrix $\WU \in \R^{\dvocab \times \dmodel}$:
\begin{equation}
  \logits = \h^{(\nlayer)} \WU^\top .
\end{equation}

Embedding and unembedding weight tying is a standard language-model design choice~\cite{pressUsingOutputEmbedding2017,inanTyingWordVectors2017}. In the \gls{gpt2}-small checkpoint used here~\cite{radford2019language}, the embedding and unembedding matrices are tied ($\WU = \WE$). In the Llama-3.1-8B checkpoint used here~\cite{grattafioriLlama3Herd2024a}, they are not. For models with untied input embeddings and output unembeddings, the choice of anchor matrix is not unique. Input embeddings provide vocabulary-indexed residual directions near the model input, whereas unembedding directions may be more appropriate for later layers. In this work we focus on input-embedding anchors and treat unembedding-based anchors as a possible extension.

\subsection{Vocabulary-Aligned Probing}

A common diagnostic for Transformer residual streams is to project intermediate post-residual streams into vocabulary-logit space. In our auxiliary \gls{logitlens}-style readout, matching the implementation used for the reported readout metric, we directly project the post-residual stream from intermediate layers through the unembedding matrix:
\begin{equation}
  \logits^{(\layer)} = \h^{(\layer)} \WU^\top .
\end{equation}

This measures which token logits are linearly decodable from an intermediate post-residual stream under this direct readout. It is separate from the \gls{ce} recovery metric, where the patched residual stream is passed through the remaining model computation before final logits are read. This background motivates the reference frame used by \gls{vasae}: vocabulary-indexed directions can be used to relate residual-stream geometry to token space. Unlike a logit-lens readout, \gls{vasae} does not decode activations into token predictions; it softly anchors learned \gls{sae} decoder directions to vocabulary-indexed reference directions.

\subsection{SAEs for Residual-Stream Analysis}

\Glspl{sae} map a post-residual stream to a sparse code and then decode the sparse code back into residual-stream space. Panel B of \figref{fig:arch} shows this encoder--decoder structure. \Gls{sae} variants differ in how they parameterize the encoder and enforce sparsity. This paper uses a \topk{} \gls{sae}~\cite{makhzaniKSparseAutoencoders2014,gaoScalingEvaluatingSparse2025}, where sparsity is achieved by keeping only the largest $k$ encoder coordinates after ReLU.

The encoder uses $\Wenc\in\R^{\dmodel\times\dsparse}$ and $\benc\in\R^{\dsparse}$, applying a learned affine map followed by ReLU and \topk{} selection:
\begin{equation}
  \label{eq:sae-sparse-encoder}
  \saecode = \enc(\h)
  = \TopK_k\!\left(\relu(\h \Wenc + \benc)\right).
\end{equation}

Here $\enc$ denotes the full sparse encoder, including the ReLU nonlinearity and \topk{} selection. The vector $\saecode$ is the sparse code. Each scalar $\saecode_i$ is one element in the sparse code vector and is the coefficient for feature $i$. The operator $\TopK_k$ keeps the largest $k$ coordinates and sets all others to zero, so each sparse code satisfies $\norm{\saecode}_0\le k$, where $\norm{\cdot}_0$ denotes the number of nonzero coordinates.

The decoder uses $\Wdec\in\R^{\dmodel\times \dsparse}$ and $\bdec\in\R^{\dmodel}$ to map the sparse code back to the residual-stream space:
\begin{equation}
  \label{eq:sae-generic-reconstruction}
  \saerecon = \saecode \Wdec^\top + \bdec .
\end{equation}
The online experiments use the extracted post-residual streams directly; we do not subtract a separate stored activation mean before the encoder. Constant offsets can be represented by the learned affine encoder and decoder bias terms.

For a training set of $N$ post-residual streams $\{\h_i\}_{i=1}^{N}$, where $N$ counts evaluated token positions, the \gls{sae} training loss is the empirical reconstruction error under this sparse encoder:
\begin{equation}
  \label{eq:sae-generic-loss}
  \loss_{\text{recon}}
  = \frac{1}{N}\sum_{i=1}^{N}\norm{\h_i - \tilde{\h}_i}_2^2 .
\end{equation}

Sparsity means only a few elements of the sparse code are nonzero for a given post-residual stream, yielding a decomposition with a small active feature set. In \secref{sec:method}, we keep the same reconstruction problem with sparse codes and add vocabulary-aligned anchoring to the dictionary.

\section{Method}
\label{sec:method}

\begin{figure}[htbp]
    \centering
    \includegraphics[width=1\linewidth]{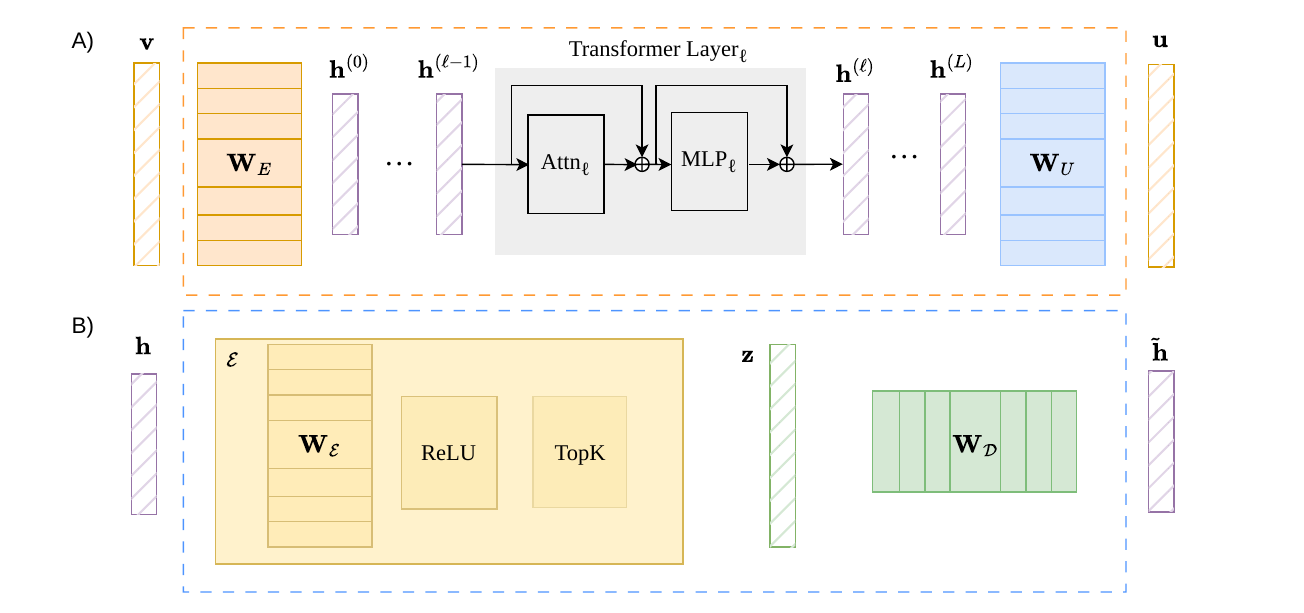}
    \caption{Transformer residual-stream decomposition with \gls{vasae}. The model learns the dictionary $\Wdec$ under a reconstruction objective plus the vocabulary-anchor loss $\loss_{\text{anchor}}$.}
    \label{fig:arch}
\end{figure}

We decompose the post-residual stream with a learned dictionary whose features are anchored to the fixed token-embedding directions by a vocabulary-aligned objective. Sparsity enters through the sparse code: for a given post-residual stream, only a few dictionary features have nonzero corresponding elements in the sparse code. Our proposed \gls{vasae} method keeps the \gls{sae} dictionary learnable and uses token embeddings only as naming anchors. The method is summarized in \figref{fig:arch}.

\subsection{\texorpdfstring{\gls{vasae}}{VASAE} Architecture}
\label{subsec:method_vasae}

For each model layer, we train a separate \gls{vasae} model on the post-residual stream from that layer. The input is $\h\in\R^{\dmodel}$, and the encoder is the same linear ReLU \topk{} map as in \eqnref{eq:sae-sparse-encoder}. This produces a nonnegative sparse code $\saecode\in\R^{\dsparse}$ with at most $k$ nonzero elements. The reconstruction is produced by the learned decoder $\Wdec$ as in \eqnref{eq:sae-generic-reconstruction}.

We write the dictionary as $\Wdec=(\mathbf{d}_1,\ldots,\mathbf{d}_{\dsparse})$, where $\mathbf{d}_i\in\R^{\dmodel}$ is feature $i$. The scalar $\saecode_i$ is one element in the sparse code vector and is the coefficient for feature $i$.

The sparse code dimension $\dsparse$ is a configurable \gls{vasae} hyperparameter. In the experiments, we instantiate the vocabulary-sized case $\dsparse=\dvocab$, so the learned dictionary has one feature slot per vocabulary item. Other dictionary sizes are compatible with the method. The decoder columns remain learnable, and token embeddings are fixed reference directions during \gls{vasae} training.
We use vocabulary-sized dictionaries in the main experiments to make the learned feature set and vocabulary anchor set comparable in scale.

\subsection{Vocabulary-Aligned Anchoring}
\label{subsec:method_vocab_anchor}

\gls{vasae} assumes that \gls{sae} decoder directions and the chosen vocabulary anchor directions live in the same dimensional vector space. For residual-stream \glspl{sae}, this assumption holds when the model's token embedding vectors have the same width as the residual stream. Architectures with factorized embeddings or mismatched dimensions would require an explicit learnable projection matrix that maps token embeddings to the residual-stream dimension, or an anchor set already defined in the residual-stream space.

For each learned feature $\mathbf{d}_i$, \gls{vasae} compares the feature with every fixed token embedding and records both the best cosine similarity and the vocabulary item that attains it:
\begin{align}
    s_i &= \max_{v\in\vocab} \cos(\mathbf{d}_i,\mathbf{w}_v), \label{eq:sim} \\
    v_i^\star &= \argmax_{v\in\vocab} \cos(\mathbf{d}_i,\mathbf{w}_v). \label{eq:token-name}
\end{align}

Thus, $s_i$ is the nearest-token alignment score: it is large only when feature $i$ lies close to some vocabulary direction. The vocabulary item $v_i^\star$ determines the intrinsic token name for the feature. The feature vector $\mathbf{d}_i$ itself remains free to move under the reconstruction objective.
Cosine similarities are computed after $L_2$-normalizing decoder columns and token embedding vectors. For each decoder column, we compute cosine similarity to all token embeddings. The nearest token embedding defines the current anchor. The loss below rewards high similarity to this anchor and is optimized together with reconstruction under the \topk{} sparse-code constraint.

To increase nearest-token alignment, we define the anchor loss so that minimizing it increases the cosine similarity between each feature and its nearest token embedding:
\begin{equation}
  \label{eq:vasae-anchor-loss}
  \loss_{\text{anchor}} = -\frac{1}{\dsparse} \sum_{i=1}^{\dsparse} s_i .
\end{equation}

Combined with reconstruction loss in \eqnref{eq:sae-generic-loss}, the total loss of \gls{vasae} is
\begin{equation}
  \loss_{\text{vasae}} =
  \loss_{\text{recon}} + \lambda_{\text{anchor}} \loss_{\text{anchor}},
\end{equation}

During optimization, $v_i^\star$ is recomputed from the current feature vector $\mathbf{d}_i$, so the nearest-token anchor can change as the dictionary changes. The token embeddings serve only as fixed reference vectors.

The anchor coefficient sets the weight of the vocabulary-alignment term relative to reconstruction. The \topk{} constraint fixes the number of active features separately. Larger $\lambda_{\text{anchor}}$ values pull decoder directions more strongly toward token embeddings, so we tune $\lambda_{\text{anchor}}$ as a regularization hyperparameter.

For large vocabularies, we compute the nearest-token scores and lookup with a memory-bounded chunked similarity implementation. The chunked computation avoids materializing the full feature-by-vocabulary similarity matrix. Details are given in \appref{apd:sim}.

\subsection{Feature Naming}
\label{subsec:method_training_naming}

After training, each feature $\mathbf{d}_i$ receives an intrinsic token name. The name is the token string corresponding to the nearest vocabulary item $v_i^\star$ in \eqnref{eq:token-name}. We use the term intrinsic token name to mean the nearest vocabulary token under cosine similarity in the chosen embedding space. This is a geometric identifier for the feature. We call this name intrinsic because it is determined by the geometry between the learned feature vector and the fixed token-embedding matrix, rather than by manually inspecting contexts where elements of the sparse code are large or by applying an additional post-hoc labeling procedure.

We report $s_i$ together with the intrinsic token name as the feature's alignment score. A larger $s_i$ means that the feature vector is closer to its named token embedding. A smaller $s_i$ means that the nearest token still gives the closest vocabulary direction, but with lower geometric support.

\section{Experiments}
\label{sec:experiments}

We evaluate whether vocabulary-aligned anchoring largely preserves \gls{sae} reconstruction, learns dictionaries with token-aligned features, and assigns token names to features with large sparse-code coefficients.

\subsection{Experimental Setup}
\label{subsec:exp_setup}

\paragraph{Dataset.}
Our training and evaluation data are post-residual-stream activations extracted from language-model forward passes on WikiText-103~\cite{merityPointerSentinelMixture2017}. We split sequences into train/validation/test subsets as follows: $50{,}000/10{,}000/5{,}000$ for GPT-2-small and $20{,}000/2{,}000/5{,}000$ for Llama-3.1-8B. Each sequence is truncated or padded to a maximum length of $128$, giving at most about $6.4$M/$1.3$M/$0.6$M token positions for GPT-2-small and $2.6$M/$0.3$M/$0.6$M token positions for Llama-3.1-8B before padding positions are masked. We run the language model on the tokenized sequences and extract post-residual-stream activations online with the \texttt{nnsight}~\cite{fiotto-kaufmanNNsightNDIFDemocratizing2025} tracing framework. We use two decoder-only Transformer checkpoints to cover a smaller GPT-2 setting and a larger  Llama setting:
\begin{itemize}
  \item \textbf{GPT-2-small}: a GPT-2 checkpoint used for full layerwise analysis. It has model dimension $\dmodel=768$, vocabulary size $\dvocab=50{,}257$, and $12$ layers~\cite{radford2019language}.
  \item \textbf{Llama-3.1-8B}: a larger open-weight Llama checkpoint used to test the method in a higher-dimensional vocabulary and residual-stream geometry. It has model dimension $\dmodel=4096$, vocabulary size $\dvocab=128{,}256$, and $32$ layers~\cite{grattafioriLlama3Herd2024a}.
\end{itemize}

\Gls{vasae} and all baselines are trained and evaluated on these extracted post-residual streams. Padding positions are excluded from training and evaluation with the tokenizer attention mask.

\paragraph{Baselines.}
\begin{itemize}
  \item \textbf{Standard \gls{sae}}: a \gls{sae} trained directly on post-residual streams without explicit anchoring to fixed token-embedding directions.
  \item \textbf{Hard-tied decoder baseline}: a comparison method that uses exact token embeddings as features. The decoder is fixed during training. Only the encoder is optimized.
\end{itemize}

The hard-tied decoder baseline tests the direct alternative of representing post-residual streams only with token-embedding features. Because its decoder features are exactly the vocabulary embeddings, it necessarily has $\dsparse = \dvocab$. It sets $\Wdec^\top = \WE$ in \eqnref{eq:sae-generic-reconstruction} and reconstructs as $\saerecon = \saecode \WE$.
The encoder is trained, but the decoder weights are frozen. This baseline tests whether hard-tying the decoder to token embeddings is sufficient.

\subsection{Reconstruction Preservation}
\label{subsec:benchmarking_vasae}

We first ask whether vocabulary-aligned anchoring changes reconstruction quality relative to a standard \gls{sae}. We report reconstruction error and cross-entropy loss after residual-stream substitution.

\paragraph{Training protocol.}
All three models, standard \gls{sae}, hard-tied \gls{sae}, and \gls{vasae}, use vocabulary-sized sparse code dimension $|\vocab|$, \topk{} sparsity with $k=32$, nonnegative sparse code, a linear encoder, and Adam with learning rate $10^{-3}$. For standard \gls{sae} and \gls{vasae}, the vocabulary-sized dimension is chosen to match the hard-tied baseline. Training runs for at most $20$ epochs with patience-$3$ early stopping on validation reconstruction loss, and we evaluate the best checkpoint on the test split. For \gls{vasae}, the anchor term is optimized during training but is not included in validation checkpoint selection. The default anchor coefficient is $\lambda_{\text{anchor}}=10^{-4}$. For Llama-3.1-8B geometric-alignment and case-study diagnostics, we additionally use $\lambda_{\text{anchor}}=5\times10^{-3}$; the supporting anchor-strength sweep is reported in \appref{apd:anchor-sensitivity}. GPT-2-small runs use float32 with batch size $32$. Llama-3.1-8B runs use bfloat16 with batch size $8$; for Llama \gls{vasae} runs, the anchor term is evaluated once every $50$ optimizer steps, and skipped-anchor steps optimize only the reconstruction loss. Each reported training run uses a single NVIDIA GH200 GPU with 96GB GPU memory. Nearest-token lookup is computed over the model tokenizer's full vocabulary. Padding positions are excluded from activation training and evaluation through the attention mask.

\begin{table*}[t]
  \centering
  \small
  \caption{Results, reported as mean $\pm$ standard deviation across layers. Arrows indicate the preferred direction for each metric. Shaded rows denote the proposed method. Best values within each model block are bolded.}
  \label{tab:vasae_benchmark}
  \begin{tabular}{@{}llcccc@{}}
    \toprule
    Model & Method & MSE $\downarrow$ & VE $\uparrow$ & CE loss $\downarrow$ & CE rec. $\uparrow$ \\
    \midrule
    GPT-2-small & Standard \gls{sae} & $\boldsymbol{2.04 \pm 2.78}$ & $\boldsymbol{0.965 \pm 0.053}$ & $4.18 \pm 0.11$ & $\boldsymbol{0.975 \pm 0.028}$ \\
    GPT-2-small & Hard-tied decoder & $110.39 \pm 62.53$ & $-0.434 \pm 1.046$ & $14.64 \pm 10.65$ & $-0.606 \pm 2.642$ \\
    \rowcolor{black!6}
    GPT-2-small & \textbf{\gls{vasae}} & $\boldsymbol{2.04 \pm 2.78}$ & $0.965 \pm 0.054$ & $\boldsymbol{4.17 \pm 0.11}$ & $\boldsymbol{0.975 \pm 0.028}$ \\
    \midrule
    Llama-3.1-8B & Standard \gls{sae} & $\boldsymbol{0.057 \pm 0.096}$ & $\boldsymbol{0.931 \pm 0.091}$ & $3.55 \pm 0.68$ & $0.906 \pm 0.075$ \\
    Llama-3.1-8B & Hard-tied decoder & $0.672 \pm 0.236$ & $-0.021 \pm 0.058$ & $11.46 \pm 0.92$ & $0.032 \pm 0.101$ \\
    \rowcolor{black!6}
    Llama-3.1-8B & \textbf{\gls{vasae}} & $0.058 \pm 0.097$ & $\boldsymbol{0.931 \pm 0.091}$ & $\boldsymbol{3.55 \pm 0.67}$ & $\boldsymbol{0.906 \pm 0.074}$ \\
    \bottomrule
  \end{tabular}
\end{table*}

\paragraph{Evaluation metrics.}
\tabref{tab:vasae_benchmark} reports four metrics. Exact formulas are given in \appref{apd:evaluation-metrics}.
\begin{itemize}
  \item \textbf{Mean Squared Error (MSE)}: residual-stream reconstruction error. Lower values mean the reconstruction is closer to the target activation.
  \item \textbf{Variance Explained (VE)}: scale-normalized reconstruction quality. Higher values mean the reconstruction preserves more target activation variance.
  \item \textbf{Cross-Entropy Loss (CE loss)}: downstream next-token loss after substituting the reconstruction at the evaluated residual-stream site. Lower values mean better preservation of model predictions.
  \item \textbf{Cross-Entropy Recovery (CE rec.)}: downstream loss recovery relative to the gap between the original residual stream and a zero-vector control. Values near $1$ match the original stream, values near $0$ match the zero-vector control, and negative values are worse than that control.
\end{itemize}
\tabref{tab:vasae_benchmark} reports mean $\pm$ standard deviation across layers. The standard deviations are across layers rather than random seeds. \gls{vasae} has reconstruction metrics comparable to a standard \gls{sae} in both tested model families. On GPT-2-small, the two learned-dictionary methods have the same \gls{ve} ($0.965$) and \gls{ce} recovery ($0.975$). On Llama-3.1-8B, they also have the same \gls{ve} ($0.931$) and \gls{ce} recovery ($0.906$). The hard-tied decoder baseline performs much worse, showing that directly forcing every feature to be an exact token embedding is not sufficient. This contrast supports the central design choice in \gls{vasae}: keep the dictionary learnable while anchoring features to token embeddings. Layerwise \gls{ve}, \gls{ce} recovery, and auxiliary \gls{logitlens} diagnostics are shown in \appref{apd:benchmark-layerwise}.

\subsection{Geometric Token Alignment}
\label{subsec:alignment_stability}

Having evaluated reconstruction, we next measure whether learned features become geometrically token aligned using the alignment score defined in \eqnref{eq:sim}. This section checks whether the anchor loss produces token-aligned decoder directions.

\begin{figure}[htbp]
  \centering
  \includegraphics[width=\linewidth]{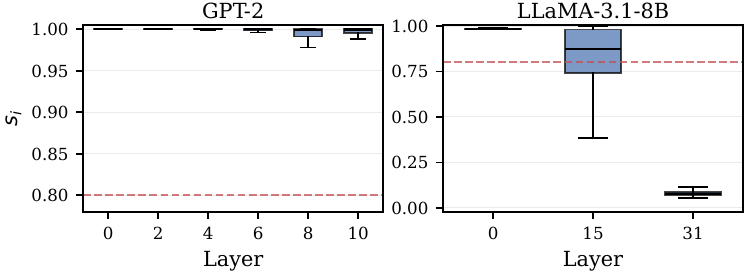}
  \caption{Geometric alignment diagnostics. Left: GPT-2-small alignment score distributions for representative even-numbered layers from $L0$ to $L10$. Right: Llama-3.1-8B alignment score distributions at $\lambda_{\text{anchor}}=5\times10^{-3}$ for $L0$, $L15$, and $L31$. The dashed line marks the diagnostic strong-alignment threshold $s_i=0.8$.}
  \label{fig:alignment_geom}
\end{figure}

\paragraph{Evaluation setup.}
For GPT-2-small, we compare \gls{vasae} with a standard \gls{sae} using $\lambda_{\text{anchor}}=10^{-4}$. For Llama-3.1-8B, we use $\lambda_{\text{anchor}}=5\times10^{-3}$ because $\lambda_{\text{anchor}}=10^{-4}$ preserves reconstruction but produces lower token alignment scores; the appendix sweep shows that $5\times10^{-3}$ changes \gls{ve} and \gls{ce} recovery only slightly on representative layers. For both model families, we use the alignment score $s_i$ and its associated nearest vocabulary item $v_i^\star$, defined in \eqnref{eq:sim} and \eqnref{eq:token-name}, respectively. With threshold $\tau=0.8$, the strongly aligned feature-index set is $\mathcal{A}_\tau=\{i:s_i\ge\tau\}$. This value was chosen from the observed separation in alignment-score distributions: standard \gls{sae} features concentrate well below $0.3$, while many \gls{vasae} features cluster near $1.0$, as shown in \figref{fig:alignment_geom}. The threshold is a diagnostic cutoff for geometric alignment; the full score distribution remains the primary evidence. We report the geometric feature-alignment rate $|\mathcal{A}_\tau|/\dsparse$ and vocabulary coverage $|\{v_i^\star:i\in\mathcal{A}_\tau\}|/\dvocab$, since many features can be aligned to the same token. Coverage is the fraction of vocabulary tokens used as nearest-token names by at least one strongly aligned feature. The denominator is the full-learned dictionary, so this rate should be read as a geometric property of the feature set.

\paragraph{GPT-2-small shows high alignment across most layers.}
For GPT-2-small, \figref{fig:alignment_geom} left summarizes representative even-numbered layers from $L0$ to $L10$. The alignment distributions are clearly separated: about $93\%$ of \gls{vasae} features in shallow and middle layers have alignment scores near $1.0$, whereas standard \gls{sae} features remain concentrated around $0.1$--$0.2$ and never exceed the diagnostic cutoff. Under this criterion, post-hoc nearest-token naming of a standard \gls{sae} is a low-alignment baseline: the nearest token exists by definition, but it is typically far from the learned feature direction. Across layers, the geometric feature-alignment rate stays between $89\%$ and $94\%$ for $L0$--$L10$ and is $68.5\%$ in $L11$. Coverage of unique aligned tokens is around $53\%$--$56\%$, indicating that multiple features can share an intrinsic token name. At this coverage level, the naming map is many-to-one: many aligned features share nearest-token names. These results support the main alignment claim: in GPT-2-small, vocabulary-aligned anchoring converts a standard \gls{sae} dictionary into a geometrically token-aligned dictionary with comparable reconstruction metrics.

\paragraph{Llama-3.1-8B alignment is shallower.}
At $\lambda_{\text{anchor}}=5\times10^{-3}$, the right panel of \figref{fig:alignment_geom} shows a depth-dependent pattern: alignment is highest at $L0$, lower at $L15$, and not comparable at $L31$. The $L0$ dictionary has $119{,}016$ strongly aligned features out of $128{,}256$ ($92.8\%$), with $77{,}371$ unique aligned token names ($60.33\%$ vocabulary coverage). These Llama-3.1-8B results indicate higher alignment in shallow layers and less stable alignment in deeper layers, matching the hypothesis that vocabulary anchoring is more compatible with residual states closer to embedding space.

\subsection{Case Studies of Intrinsic Token Names}
\label{subsec:case_studies}

\paragraph{Case-study setup.}
We next inspect the intrinsic token names assigned to aligned features that are active in context. These case studies show token-level intrinsic names on individual prompts. The main location example uses the fixed input sentence ``The cafe is located on Baker Street, just around the corner from the avenue''. We tokenize this sentence with each model's tokenizer and display the first 12 non-padding token positions. The GPT-2-small visualization uses representative even-numbered layers from $L0$ to $L10$; the Llama-3.1-8B visualization uses layers $L0,L15,L31$. For each displayed layer $\ell$, let $\mathbf{z}_p\in\R^{\dsparse}$ be the sparse code coefficient at token position $p$ in a sentence of length $T$, and let $\mathcal{A}_{\tau}$ be the strongly aligned feature-index set from the previous subsection. If a feature is active at many positions in the same sentence, raw activation alone can choose that same feature repeatedly for display. We therefore choose the aligned feature that is most elevated relative to the sentence average:
\begin{equation}
  \bar{\mathbf{z}}=\frac{1}{T}\sum_{q=1}^{T}\mathbf{z}_q,
  \qquad
  i_p^\star\in\argmax_{i\in\mathcal{A}_{\tau}}
  [\mathbf{z}_p-\bar{\mathbf{z}}]_i .
\end{equation}
We display the intrinsic token name of the chosen feature, namely the token string corresponding to $v_{i_p^\star}^\star$. The relative sparse code is used only to choose the displayed feature. The color shows the raw sparse-code activation $[\mathbf{z}_p]_{i_p^\star}$.

\begin{figure}[htbp]
  \centering
  \includegraphics[width=0.96\linewidth]{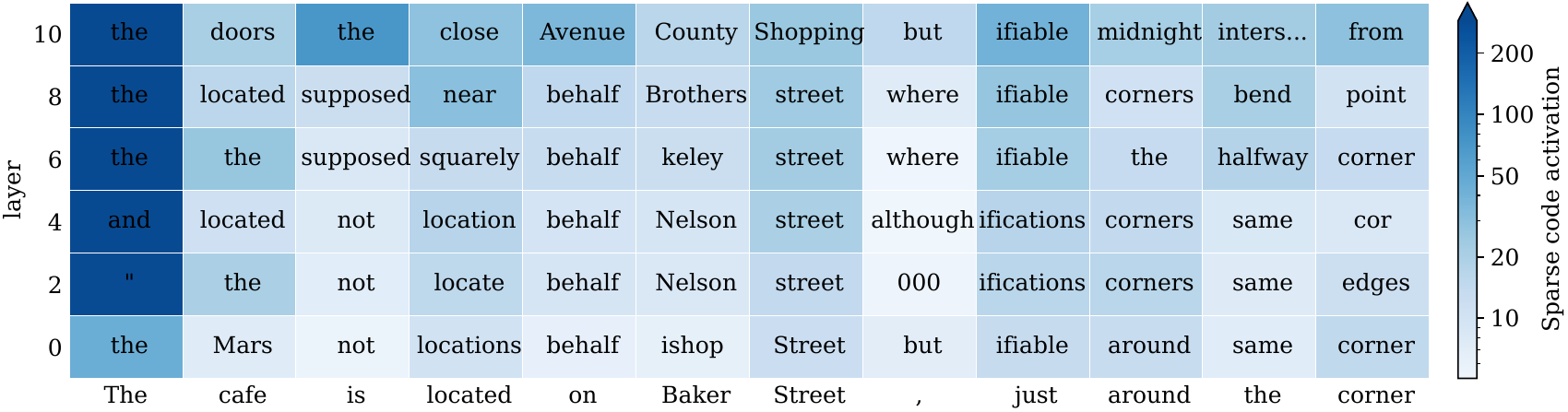}
  \caption{
    GPT-2-small location case study. Each cell shows the intrinsic token name of the aligned feature with the largest sentence-centered sparse-code value at a layer and input position. Color indicates the raw sparse-code activation of that feature. Shallow and middle layers show location-related intrinsic token names.
  }
  \label{fig:gpt2_case_studies_main}
\end{figure}

\paragraph{GPT-2-small example.}
\figref{fig:gpt2_case_studies_main} shows a representative GPT-2-small location example. Shallow and middle layers repeatedly display location-related intrinsic token names such as \textit{located}, \textit{location}, \textit{Street}, \textit{around}, \textit{corner}, and \textit{corners} around the input phrase ``Baker Street'' and the surrounding spatial context. Additional examples are shown in \appref{apd:additional-case-studies}, including adjective/adverb words, award-related words, and self-introduction words. These visualizations show intrinsic token names after geometric alignment has been established and provide qualitative inspection examples.

\begin{figure}[htbp]
  \centering
  \includegraphics[width=0.96\linewidth]{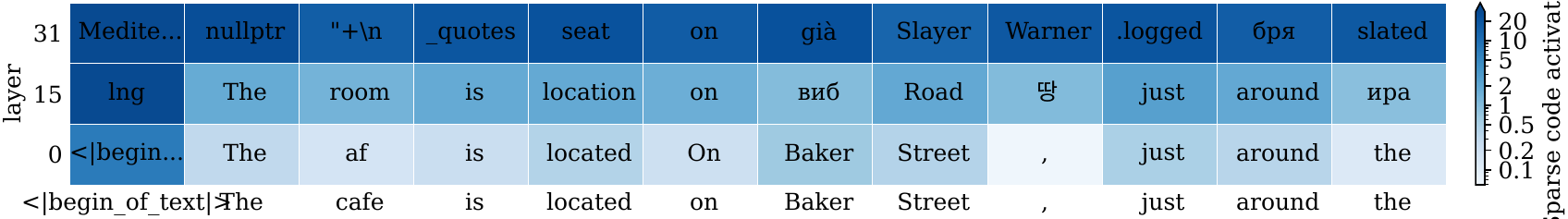}
  \caption{
    Llama-3.1-8B case study at $\lambda_{\text{anchor}}=5\times10^{-3}$. Each cell shows the intrinsic token name of the feature chosen by the same sentence-centered sparse-code rule as \figref{fig:gpt2_case_studies_main}. Color indicates the raw sparse-code activation of that feature. The shallow representative layer shows location-related token names, the middle representative layer is less consistent, and the final representative layer is unstable.
  }
  \label{fig:llama_case_study_main}
\end{figure}

\paragraph{Llama-3.1-8B example.}
\figref{fig:llama_case_study_main} applies the same sentence-centered rule to Llama-3.1-8B at $\lambda_{\text{anchor}}=5\times10^{-3}$ for the representative layers $L0$, $L15$, and $L31$. The shallow layer shows location-related token names, the middle layer is less consistent, and the final layer is mostly unstable. This qualitative pattern matches the geometric diagnostics in \figref{fig:alignment_geom}.

\section{\texorpdfstring{Discussion}{Discussion}}
\label{sec:discussion}

\paragraph{What VASAE names do and do not mean.}
A \gls{vasae} name is a vocabulary-indexed handle for a learned decoder direction: the token whose embedding is nearest to that direction. Its role is to attach a model-internal token reference to the feature during training, rather than adding a separate naming step after feature learning.

The main point to avoid over-reading is that this token handle is not a feature explanation. It gives a starting point for inspection, but the meaning or role of the feature still has to be established from how it activates, how it connects to other model components, or what changes under intervention.

\paragraph{What counts as explaining a feature.}
Feature interpretation depends on the research goal. The examples below illustrate a distinction that is easy to blur in \gls{sae} analysis: a feature can be useful because information is readable from it, because it participates in an internal computation, or because intervening on it changes behavior. Each use calls for its own evidence. This matters because a named feature can make an interpretation look more complete than it is. A safer use of feature names is to state which question the feature is being used to answer and what evidence supports that question.

For example, \textbf{semantic readout} treats a feature as a direction from which some information may be readable. A feature whose nearest token is location-related may be a candidate direction for residual-stream information about place names, street names, spatial prepositions, or local descriptions of where something is; evidence for this use comes from top-activating contexts, activation distributions across prompts, and comparisons with nearby candidate features. \textbf{Computational role} asks whether the feature participates in an internal calculation, such as tracking an intermediate variable for entity resolution, maintaining a syntactic constraint, combining local context into a next-token preference, or passing information from one component to another; this requires circuit-level evidence such as ablations, activation patching, path patching, or dependencies between upstream and downstream features and modules. \textbf{Behavioral control} asks whether intervening on the feature changes outputs in a predictable way, for example shifting generated text toward a more formal, cautious, specific, or creative style. In that setting, the interpretation is tied to the intervention effect, not only to the contexts where the feature activates or to its nearest-token name.

For the goal of understanding how a language model implements a behavior, this suggests shifting attention from whether an isolated feature has the right name to how named features enter circuits. The central object is the mechanism: how features, attention heads, MLP components, and residual-stream directions pass information forward, transform it, and affect behavior. In this view, feature naming is a preliminary indexing tool. The analysis should ask which components use the feature, which downstream paths change when it is perturbed, and which behavior is explained by the resulting circuit.

\section{Related Work}

\subsection{Vocabulary Readout and Representation Geometry}

Probing work formalizes how to test whether intermediate post-residual streams linearly encode linguistic or task-relevant attributes~\cite{tenneyBERTRediscoversClassical2019a,hewittDesigningInterpretingProbes2019a}. In autoregressive Transformers, layerwise vocabulary-readout methods, from \gls{logitlens}-style readouts to tuned-lens variants, show how token-level predictions can be decoded and analyzed across Transformer layers~\cite{belroseElicitingLatentPredictions2025}. Our work is complementary to this literature: rather than studying decodability alone, we study decompositions of the residual stream itself.

Related work on representation geometry studies how information is arranged in language-model vector spaces, including whether representations are anisotropic or dominated by common directions~\cite{ethayarajhHowContextualAre2019a,muAllbuttheTopSimpleEffective2018a}. These works motivate geometric analysis of language-model representations. We use token embeddings as the reference frame because they provide vocabulary-indexed directions in the residual-stream space, while weight-tying work shows that embedding and output-token geometry are central interfaces in language models~\cite{pressUsingOutputEmbedding2017,inanTyingWordVectors2017}. Our question is more specific: can a learned dictionary stay close enough to vocabulary directions to provide intrinsic token names while still reconstructing the post-residual stream from sparse codes, without being forced into the exact token-embedding directions? This motivates our anchor choice: when token embeddings share dimensionality with the residual-stream \gls{sae} decoder, they define vocabulary-indexed directions against which learned decoder columns can be compared.

\subsection{Sparse Codes and Mechanistic Analysis}

Superposition motivates sparse feature dictionaries as a way to separate features that share model dimensions~\cite{elhageToyModelsSuperposition2022a}. Dictionary-learning work decomposes language-model activations into sparse features that are more interpretable than individual neurons~\cite{bricken2023monosemanticity,hubenSparseAutoencodersFind2024}. Subsequent work scales and evaluates \glspl{sae} across language-model settings~\cite{gaoScalingEvaluatingSparse2025,DBLP:conf/icml/KarvonenRLTBCLF25}, and sparse feature circuits use these features for causal graph analysis and editing~\cite{marksSparseFeatureCircuits2025}. Existing interpretation workflows usually inspect top-activating contexts or apply automated natural-language explanations after training~\cite{bills2023language,pauloAutomaticallyInterpretingMillions2025}. Mechanistic interpretability work more broadly studies how Transformer computations factor into internal mechanisms, including module-level accounts such as feed-forward key--value memory~\cite{gevaTransformerFeedForwardLayers2021a}. Our contribution keeps the \gls{sae} reconstruction objective and adds vocabulary-aligned anchoring, so learned features can receive candidate intrinsic token names. The main claim is reconstruction-preserving geometric token alignment.

\section{Conclusion}

We presented \gls{vasae}, a residual-stream \gls{sae} whose learned dictionary is vocabulary-aligned to token embeddings. The difference from a standard \gls{sae} is the anchor term: both models learn sparse codes to reconstruct the post-residual stream, but \gls{vasae} also encourages each feature to remain close to a nearest fixed token-embedding direction, giving the feature an intrinsic token name. This added constraint yields reconstruction metrics comparable to a standard \gls{sae} in the tested GPT-2-small and Llama-3.1-8B settings, while producing strongly token-aligned features in GPT-2-small and in shallow- and middle-layer Llama settings at $\lambda_{\text{anchor}}=5\times10^{-3}$. Case studies show intrinsic token names in shallow and middle GPT-2 layers, including location words, award-related words, self-introduction words, and adjective/adverb words. Llama-3.1-8B also shows the main boundary condition: token alignment depends on anchor coefficient and remains unstable in the final layer under the tested settings. Overall, \gls{vasae} keeps the sparse reconstruction role of a standard \gls{sae} but adds geometrically supported nearest-token names for features.

\paragraph{Limitations.}
Our analysis is limited to residual-stream \glspl{sae} on two open-weight language-model families. The main alignment evidence is highest for GPT-2-small. In Llama-3.1-8B, stable alignment appears under the larger tested anchor coefficient and in the tested shallow and middle representative layers, while the final layer remains unstable. We anchor only to the input embedding matrix in this work; for untied models, testing anchors based on the unembedding matrix is left open. Broader anchor-strength sweeps, additional model families, additional model scales, and additional initialization baselines would further test the scope of the geometric alignment result. The reported evidence supports geometric alignment, not causal interpretation of the token names.

\bibliographystyle{icml2026}
\bibliography{ref}


\appendix

\section{Evaluation Metrics} \label{apd:evaluation-metrics}

This appendix collects the evaluation and diagnostic metrics used in the main text and sensitivity-analysis tables.

\paragraph{MSE.}
Given evaluation post-residual streams $\{\h_i\}_{i=1}^{M}$ and reconstructions $\{\tilde{\h}_i\}_{i=1}^{M}$, where $M$ counts valid token positions after flattening token positions across sequences, we define \gls{mse} as
\begin{equation}
  \label{eq:eval-mse}
  \mathrm{MSE} = \frac{1}{M\dmodel} \sum_{i=1}^{M} \norm{\h_i - \tilde{\h}_i}_2^2.
\end{equation}

\paragraph{VE.}
Let
\begin{equation}
  \label{eq:eval-mean}
  \bar{\h} = \frac{1}{M} \sum_{i=1}^{M} \h_i
\end{equation}
denote the empirical mean target post-residual stream. The target variance is
\begin{equation}
  \label{eq:eval-var}
  \mathrm{Var} = \frac{1}{M\dmodel} \sum_{i=1}^{M} \norm{\h_i - \bar{\h}}_2^2.
\end{equation}
Using \gls{mse} and this variance, \gls{ve} is
\begin{equation}
  \label{eq:main-ve}
  \mathrm{VE}
  = 1 - \frac{\sum_{i=1}^{M} \norm{\h_i - \tilde{\h}_i}_2^2}{\sum_{i=1}^{M} \norm{\h_i - \bar{\h}}_2^2}
  = 1 - \frac{\mathrm{MSE}}{\mathrm{Var}}.
\end{equation}
Higher \gls{ve} means more variance is recovered.

\paragraph{CE loss.}
For the \gls{ce} reconstruction check, we evaluate three residual-stream substitutions at the evaluated site: the \gls{sae}-style reconstruction $\tilde{\h}$, the original $\h$, and the zero vector $\mathbf{0}$. For substitution $r\in\{\mathrm{SAE},\mathrm{Id},0\}$, let $\probv_i^{(r)}$ be the resulting next-token distribution at evaluated position $i$, and let $y_i$ be the target next token. The average next-token cross-entropy is
\begin{equation}
  \CE_r = -\frac{1}{M}\sum_{i=1}^{M}\log \probv_i^{(r)}(y_i).
\end{equation}
Here $r=\mathrm{SAE}$ uses $\tilde{\h}$, $r=\mathrm{Id}$ uses the original $\h$, and $r=0$ uses $\mathbf{0}$. The \gls{ce} loss reported in \tabref{tab:vasae_benchmark} is $\CE_{\mathrm{SAE}}$.

\paragraph{CE rec.}
We compute \gls{ce} recovery as
\begin{equation}
  \label{eq:main-ce-recovery}
  \mathrm{CERec} =
  1 - \frac{\CE_{\mathrm{SAE}} - \CE_{\mathrm{Id}}}{\CE_0 - \CE_{\mathrm{Id}}}.
\end{equation}
A value of $0$ corresponds to the zero-vector control. A value of $1$ corresponds to the original residual stream under this normalization.
Equivalently, \gls{ce} recovery measures how much of the \gls{ce} gap between the zero-vector control and the original residual stream is recovered by the \gls{sae}-style reconstruction: values near $1$ preserve the original next-token loss, values near $0$ match the zero-vector control, and negative values are worse than that control under this normalization.

\paragraph{\gls{logitlens} accuracy.}
As an auxiliary direct-readout diagnostic, we compare the top-1 token obtained from the original post-residual stream with the top-1 token obtained from its reconstruction after projection through the unembedding matrix:
\begin{equation}
  \begin{aligned}
    \mathrm{LogitlensAcc}
    &= \frac{1}{M}\sum_{i=1}^{M}
    \mathbf{1}\!\left[a_i=\tilde{a}_i\right],\\
    a_i
    &= \argmax_{v\in\vocab} \h_i \mathbf{u}_v^\top,\\
    \tilde{a}_i
    &= \argmax_{v\in\vocab} \tilde{\h}_i \mathbf{u}_v^\top .
  \end{aligned}
\end{equation}
Here $a_i$ and $\tilde{a}_i$ are the original and reconstructed top-1 tokens under the direct readout, and $\mathbf{u}_v$ denotes the row of the unembedding matrix $\WU$ for token $v$. This metric checks agreement under a direct layerwise readout. It is separate from \gls{ce} recovery, which substitutes the reconstruction into the model and evaluates the final next-token loss after the remaining computation.

\section{Anchor-Strength Sensitivity Analysis}
\label{apd:anchor-sensitivity}

\tabref{tab:anchor_sensitivity} reports a representative sweep over $\lambda_{\text{anchor}}$ while keeping the same \topk{} \gls{sae} architecture. The setting $\lambda_{\text{anchor}}=0$ disables the anchor term and serves as the component ablation.

\begin{table*}[t]
  \centering
  \small
  \setlength{\tabcolsep}{6pt}
  \caption{Representative anchor-strength sensitivity analysis supporting the larger-coefficient Llama alignment setting. Arrows indicate the preferred direction for each metric. Best values within each model-layer block and metric column are bolded.}
  \label{tab:anchor_sensitivity}
  \begin{tabular}{@{}llcc@{\hspace{3em}}llcc@{}}
    \toprule
    \multicolumn{4}{c}{GPT-2-small} & \multicolumn{4}{c}{Llama-3.1-8B} \\
    \midrule
    Layer & $\lambda_{\text{anchor}}$ & VE $\uparrow$ & CE rec. $\uparrow$ &
    Layer & $\lambda_{\text{anchor}}$ & VE $\uparrow$ & CE rec. $\uparrow$ \\
    \midrule
    \multirow{3}{*}{L5} & $0$ & $\boldsymbol{0.9938}$ & $\boldsymbol{0.9865}$ &
    \multirow{3}{*}{L15} & $0$ & $\boldsymbol{0.9818}$ & $0.9352$ \\
    & $10^{-4}$ & $\boldsymbol{0.9938}$ & $0.9863$ &
    & $10^{-4}$ & $0.9817$ & $0.9351$ \\
    & $5\times 10^{-3}$ & $0.9937$ & $0.9854$ &
    & $5\times 10^{-3}$ & $\boldsymbol{0.9818}$ & $\boldsymbol{0.9357}$ \\
    \midrule
    \multirow{3}{*}{L11} & $0$ & $\boldsymbol{0.8155}$ & $0.8906$ &
    \multirow{3}{*}{L31} & $0$ & $0.6141$ & $0.6303$ \\
    & $10^{-4}$ & $0.8152$ & $\boldsymbol{0.8915}$ &
    & $10^{-4}$ & $0.6147$ & $\boldsymbol{0.6434}$ \\
    & $5\times 10^{-3}$ & $0.8149$ & $0.8904$ &
    & $5\times 10^{-3}$ & $\boldsymbol{0.6151}$ & $0.6391$ \\
    \bottomrule
  \end{tabular}
\end{table*}

Changing $\lambda_{\text{anchor}}$ from $0$ to $5\times10^{-3}$ changes \gls{ve} and \gls{ce} recovery only slightly on these representative GPT-2-small and Llama-3.1-8B layers. This sweep indicates empirical robustness over the tested range; $\lambda_{\text{anchor}}$ remains a tunable hyperparameter in other architectures or layers.

\section{Layerwise Diagnostics}
\label{apd:benchmark-layerwise}

\begin{figure}[H]
  \centering
  \begin{minipage}[t]{0.49\linewidth}
    \centering
    \includegraphics[width=\linewidth]{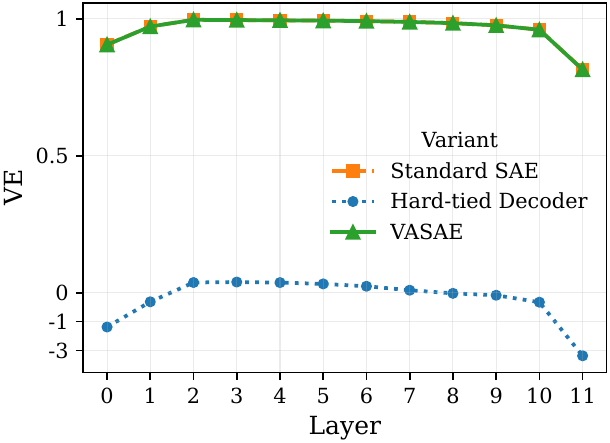}
  \end{minipage}\hfill
  \begin{minipage}[t]{0.49\linewidth}
    \centering
    \includegraphics[width=\linewidth]{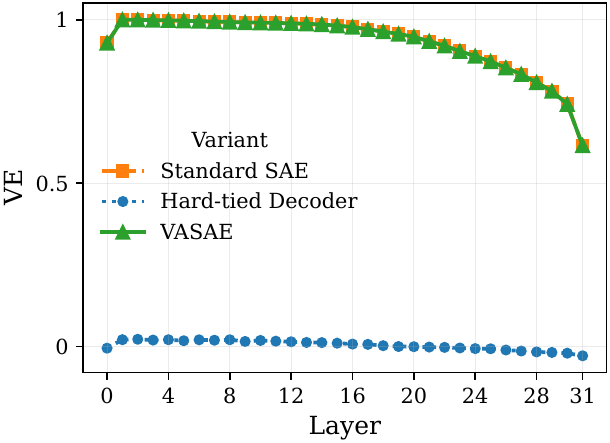}
  \end{minipage}
  
  \vspace{0.35em}

  \begin{minipage}[t]{0.49\linewidth}
    \centering
    \includegraphics[width=\linewidth]{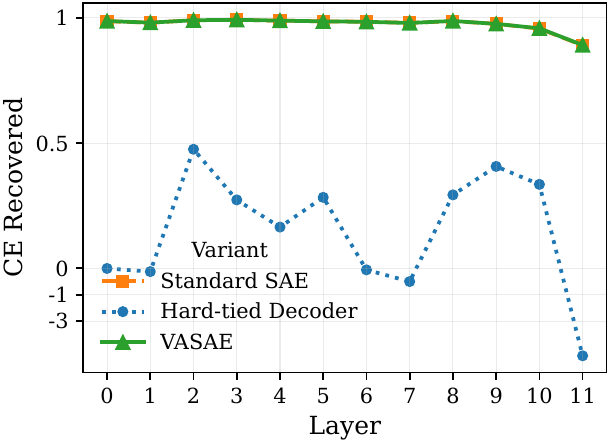}
  \end{minipage}\hfill
  \begin{minipage}[t]{0.49\linewidth}
    \centering
    \includegraphics[width=\linewidth]{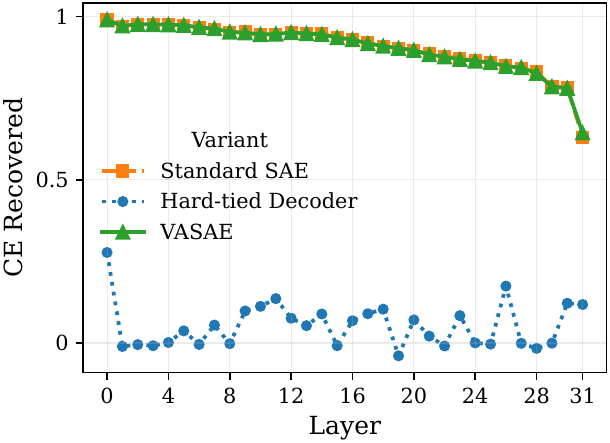}
  \end{minipage}

  \vspace{0.35em}

  \begin{minipage}[t]{0.49\linewidth}
    \centering
    \includegraphics[width=\linewidth]{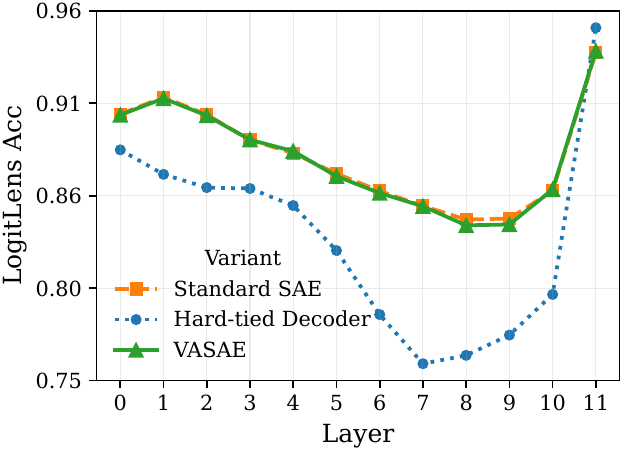}
  \end{minipage}\hfill
  \begin{minipage}[t]{0.49\linewidth}
    \centering
    \includegraphics[width=\linewidth]{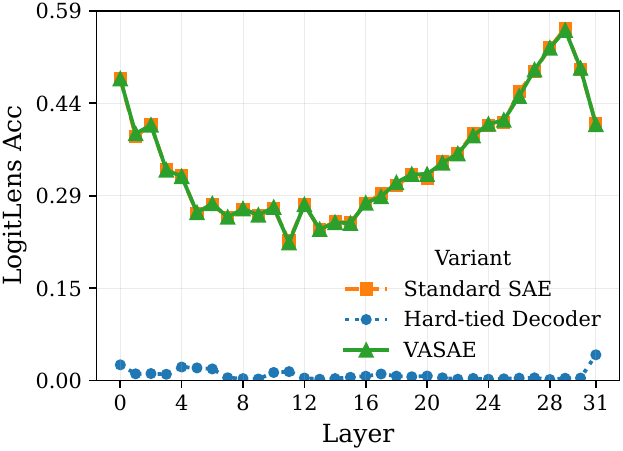}
  \end{minipage}
  \caption{Layerwise benchmark diagnostics for GPT-2-small and Llama-3.1-8B. Top: \gls{ve}. Middle: \gls{ce} recovery. Bottom: \gls{logitlens} top-1 agreement under direct unembedding.}
  \label{fig:benchmark_layerwise}
\end{figure}

The aggregate benchmark in \tabref{tab:vasae_benchmark} reports mean and standard deviation across layers. The layerwise diagnostics in \figref{fig:benchmark_layerwise} show layer-specific failures that are averaged out in aggregate means. We group the diagnostics by what they measure: \gls{ve} measures residual-space reconstruction, \gls{ce} recovery measures downstream loss after running the remaining model computation, and \gls{logitlens} accuracy measures agreement under direct unembedding.

\Gls{vasae} and the standard \gls{sae} nearly overlap in \gls{ve} and \gls{ce} recovery across layers in both model families, matching the aggregate benchmark. This supports reconstruction preservation in the tested cases; trade-offs in other settings remain empirical. The hard-tied decoder baseline is much lower on these two metrics, especially in GPT-2-small late layers and across Llama-3.1-8B. The \gls{logitlens} curves show a different direct-readout pattern: GPT-2-small hard-tying keeps higher top-1 agreement in several layers, while its \gls{ce} recovery remains lower. This indicates that direct unembedding agreement and downstream functional recovery capture different aspects of reconstruction. Llama-3.1-8B has lower \gls{logitlens} agreement overall than GPT-2-small under this direct readout.

\section{Geometric Alignment Metrics}
\label{apd:geometric-alignment-metrics}

For each feature $i$, we use the nearest-token alignment score $s_i$ and nearest vocabulary item $v_i^\star$ defined in \subsecref{subsec:method_vocab_anchor}. Let $\mathcal{A}_\tau=\{i\in\{1,\ldots,\dsparse\}:s_i\ge\tau\}$ denote the strongly aligned feature set for the diagnostic cutoff $\tau$ used in \subsecref{subsec:alignment_stability}.
Geometric alignment rate is the proportion of features that are aligned with tokens:
\begin{equation}
  \mathrm{GeomAlignRate}_\tau = \frac{|\mathcal{A}_\tau|}{\dsparse}.
\end{equation}
Vocabulary coverage is the proportion of the vocabulary that has at least one feature aligned to it:
\begin{equation}
  \mathrm{Coverage}_\tau = \frac{|\{v_i^\star:i\in\mathcal{A}_\tau\}|}{\dvocab}.
\end{equation}
The geometric alignment denominator is the full-learned feature set. 

\section{Additional Intrinsic-Name Case Studies}
\label{apd:additional-case-studies}

\begin{figure}[H]
  \centering
  \includegraphics[width=0.96\linewidth]{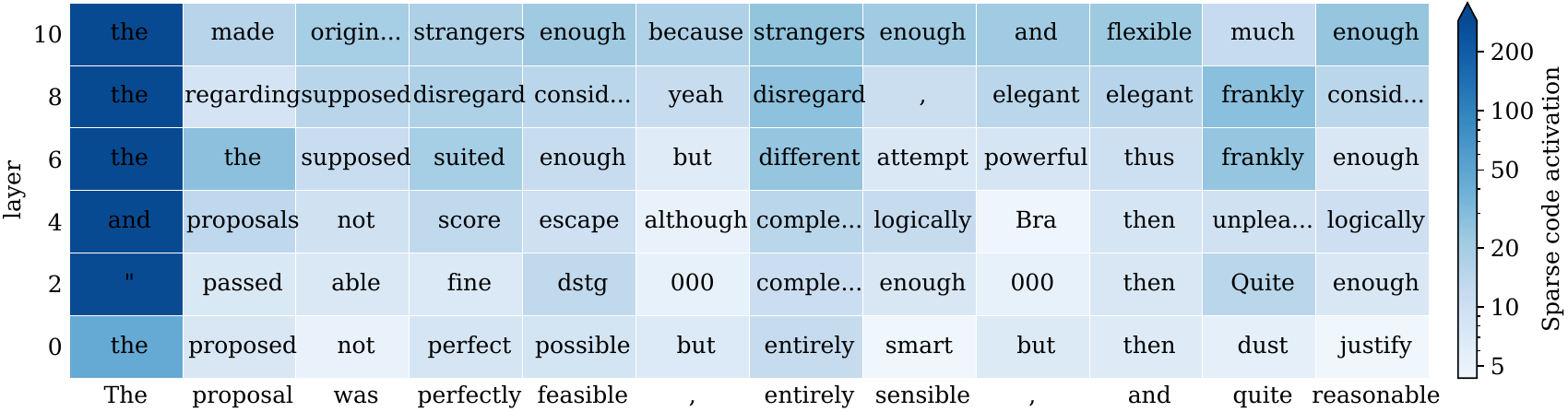}
  \caption{Additional GPT-2-small adjective/adverb example. Each cell shows the intrinsic token name of the feature chosen by the same sentence-centered sparse-code rule as \figref{fig:gpt2_case_studies_main}. Color indicates the raw sparse-code activation of that feature. Displayed names often track feasibility, degree, and evaluation words in the prompt.}
  \label{fig:apd_gpt2_case_morphology}
\end{figure}

\begin{figure}[H]
  \centering
  \includegraphics[width=0.96\linewidth]{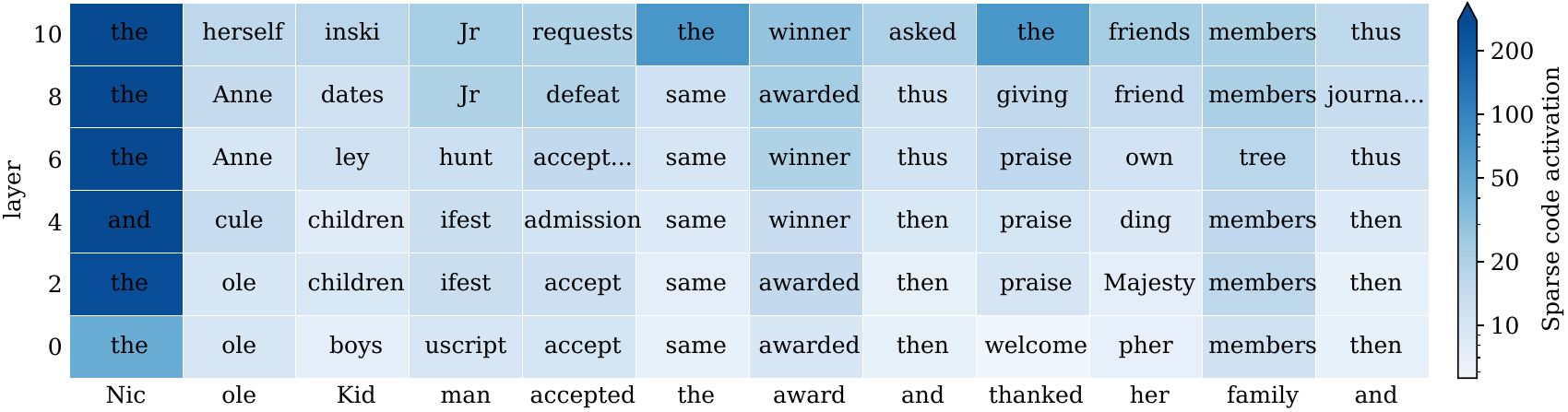}
  \vspace{0.5em}
  \includegraphics[width=0.96\linewidth]{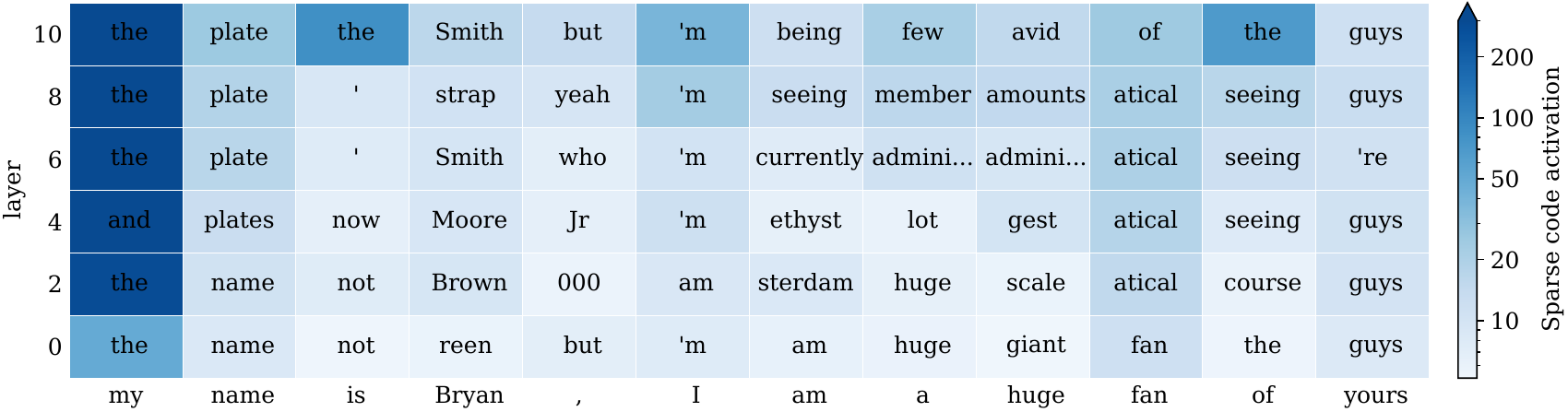}
  \caption{Additional qualitative GPT-2-small intrinsic-name examples. Each cell shows the intrinsic token name of the feature chosen by the same sentence-centered sparse-code rule as \figref{fig:gpt2_case_studies_main}. Color indicates the raw sparse-code activation of that feature. The examples show displayed names around award-related and self-introduction contexts.}
  \label{fig:apd_gpt2_case_studies}
\end{figure}

These examples extend the main GPT-2-small location case study by applying the same sentence-centered feature-display rule to other prompt types. \figref{fig:apd_gpt2_case_morphology} uses an adjective/adverb sentence and reports intrinsic token names related to feasibility, degree, and evaluation, including \textit{perfect}, \textit{possible}, \textit{entirely}, \textit{suited}, \textit{different}, \textit{enough}, \textit{completely}, \textit{logically}, \textit{elegant}, \textit{frankly}, and \textit{flexible}. \figref{fig:apd_gpt2_case_studies} covers a named-entity award sentence and a self-introduction sentence. In the Nicole Kidman sentence, reported names include award and social-context tokens such as \textit{accept}, \textit{acceptance}, \textit{awarded}, \textit{winner}, \textit{praise}, \textit{friend}, \textit{friends}, and \textit{members}. In the self-introduction sentence, reported names include personal-reference and fandom-related tokens such as \textit{name}, \textit{am}, \textit{'m}, \textit{fan}, and \textit{fans}. These examples show the same sentence-centered feature-display pattern beyond the location prompt, while remaining qualitative case studies.

\section{Implementation Details for Similarity Score} \label{apd:sim}
Computing the similarity scores $\{s_i\}_{i=1}^{\dsparse}$ defined in \eqnref{eq:sim} independently for each feature requires a nested loop over the vocabulary size $\dvocab$ and sparse code dimension $\dsparse$. To leverage the highly parallel architecture of modern GPUs, we reframe this computation into dense matrix multiplications.

The cosine similarity between a feature $\mathbf{d}_i$ and a token embedding $\mathbf{w}_v$ is equivalent to the dot product of their $L_2$-normalized vectors:
\begin{equation}
    \cos(\mathbf{d}_i,\mathbf{w}_v) = \left(\frac{\mathbf{d}_i}{\|\mathbf{d}_i\|_2}\right) \left(\frac{\mathbf{w}_v}{\|\mathbf{w}_v\|_2}\right)^\top.
\end{equation}

We perform row-wise $L_2$-normalization on $\Wdec^\top$ and $\WE$ to obtain the normalized feature-by-model matrix $\hat{\mathbf{D}}\in \mathbb{R}^{\dsparse \times \dmodel}$ and normalized embedding matrix $\hat{\WE}\in \mathbb{R}^{\dvocab \times \dmodel}$. The full pairwise similarity matrix $\mathbf{S} \in \mathbb{R}^{\dsparse \times \dvocab}$ can then be represented via a single matrix multiplication:
\begin{equation}
    \mathbf{S} = \hat{\mathbf{D}} \hat{\WE}^\top,
\end{equation}
where the element $\mathbf{S}_{i, v} = \cos(\mathbf{d}_i,\mathbf{w}_v)$. The final similarity score $s_i$ for each feature is extracted by taking the maximum value along the rows of $\mathbf{S}$:
\begin{equation}
    s_i = \max_{1 \le v \le \dvocab} \mathbf{S}_{i, v}.
\end{equation}

While mathematically straightforward, materializing the dense matrix $\mathbf{S}$ requires $\mathcal{O}(\dsparse \times \dvocab)$ memory. For typical \gls{sae} dictionary sizes and vocabulary size, this rapidly leads to Out-Of-Memory (OOM) errors. 

To resolve this, we implement a chunked computation strategy. We partition $\hat{\mathbf{D}}$ along the row dimension into contiguous blocks of size $B$. For each block $\hat{\mathbf{D}}_k \in \mathbb{R}^{B \times \dmodel}$, we compute the local similarity sub-matrix $\mathbf{S}_k = \hat{\mathbf{D}}_k \hat{\WE}^\top \in \mathbb{R}^{B \times \dvocab}$, immediately apply the row-wise maximum reduction to obtain the local scores $\mathbf{s}_k \in \mathbb{R}^B$, and discard $\mathbf{S}_k$. This strategy strictly bounds the peak memory footprint to $\mathcal{O}(B \times \dvocab)$.

The PyTorch implementation, mapping directly to our defined matrices $\Wdec$ and $\WE$, is provided in Listing \ref{lst:sim_chunk}.

\begin{lstlisting}[
style=arxivpython,
float=t,
caption={Memory-efficient PyTorch implementation for computing \eqnref{eq:sim}.},
label={lst:sim_chunk}
]
def compute_sae_similarity(
    W_dec: torch.Tensor,
    W_E: torch.Tensor,
    chunk_size: int = 2048,
) -> torch.Tensor:
    """Compute nearest-token cosine scores.
    
    Args:
        W_dec: (d_model, d_sparse) decoder weights.
        W_E: (d_vocab, d_model) token embeddings.
        chunk_size: Number of features per block.
        
    Returns:
        s: (d_sparse,) max score per feature.
    """
  
    # Normalize features and token embeddings.
    D_hat = F.normalize(W_dec.T, p=2, dim=1)
    E_hat = F.normalize(W_E.to(D_hat.dtype), p=2, dim=1)

    s_parts = []
    d_sparse = D_hat.size(0)
    
    # Chunk along features to prevent OOM.
    for i in range(0, d_sparse, chunk_size):
        # S_k shape: (current_chunk_size, d_vocab)
        S_k = D_hat[i : i + chunk_size] @ E_hat.T
        
        # Reduce immediately over tokens.
        s_k, _ = torch.max(S_k, dim=1)
        s_parts.append(s_k)

    # Concatenate chunks into the score vector.
    return torch.cat(s_parts)
\end{lstlisting}


\end{document}